\documentclass[letterpaper]{article} 
\usepackage{aaai24}  
\usepackage{times}  
\usepackage{helvet}  
\usepackage{courier}  
\usepackage[hyphens]{url}  
\usepackage{graphicx} 
\usepackage{natbib}  
\usepackage{caption} 
\usepackage{algorithm}
\usepackage{algorithmic}
\usepackage{amsmath}
\usepackage{amssymb}
\usepackage{amsthm}
\usepackage{xcolor}
\usepackage{booktabs}
\usepackage{multirow}
\usepackage{array} 
\usepackage[utf8]{inputenc}
\usepackage{color}
\usepackage{newfloat}
\usepackage{listings}
\DeclareCaptionStyle{ruled}{labelfont=normalfont,labelsep=colon,strut=off} 
\floatstyle{ruled}
\newfloat{listing}{tb}{lst}{}
\floatname{listing}{Listing}
\title{A Factuality and Diversity Reconciled Decoding Method for Knowledge-Grounded Dialogue Generation}
\author{
    Chenxu Yang\textsuperscript{\rm 1,2},  Zheng Lin\textsuperscript{\rm 1,2}\thanks{\ \ \ Zheng Lin is the corresponding author. }, Chong Tian\textsuperscript{\rm 4},  Liang Pang\textsuperscript{\rm 3},  Lanrui Wang\textsuperscript{\rm 1,2},  \\
{\bf Zhengyang Tong\textsuperscript{\rm 1,2}, Qirong Ho\textsuperscript{\rm 4}, Yanan Cao\textsuperscript{\rm 1,2}, Weiping Wang\textsuperscript{\rm 1}
}}
\affiliations{
   \textsuperscript{\rm 1}Institute of Information Engineering, Chinese Academy of Sciences, Beijing, China \\
  \textsuperscript{\rm 2}School of Cyber Security, University of Chinese Academy of Sciences, Beijing, China \\
  \textsuperscript{\rm 3}Institute of Computing Technology, CAS \quad  
  \textsuperscript{\rm 4}Mohamed bin Zayed University of AI \\
  \textrm{\{}yangchenxu,wanglanrui,linzheng,tongzhengyang\textrm{\}}@iie.ac.cn \\
  {\textrm{\{}refrainkon,hoqirong\textrm{\}}@gmail.com, pangliang@ict.ac.com}
}

\usepackage{bibentry}

\begin{document}

\maketitle

\begin{abstract}

Grounding external knowledge can enhance the factuality of responses in dialogue generation. However, excessive emphasis on it might result in the lack of engaging and diverse expressions. Through the introduction of randomness in sampling, current approaches can increase the diversity. Nevertheless, such sampling method could undermine the factuality in dialogue generation. In this study, to discover a solution for advancing creativity without relying on questionable randomness and to subtly reconcile the factuality and diversity within the source-grounded paradigm, a novel method named DoGe is proposed. DoGe can dynamically alternate between the utilization of internal parameter knowledge and external source knowledge based on the model's factual confidence. Extensive experiments on three widely-used datasets show that DoGe can not only enhance response diversity but also maintain factuality, and it significantly surpasses other various decoding strategy baselines. \footnote{The code will be released at GitHub upon publication.}
\end{abstract}

\section{Introduction}

\begin{figure}[htbp]
  \centerline{\includegraphics[scale=0.24]{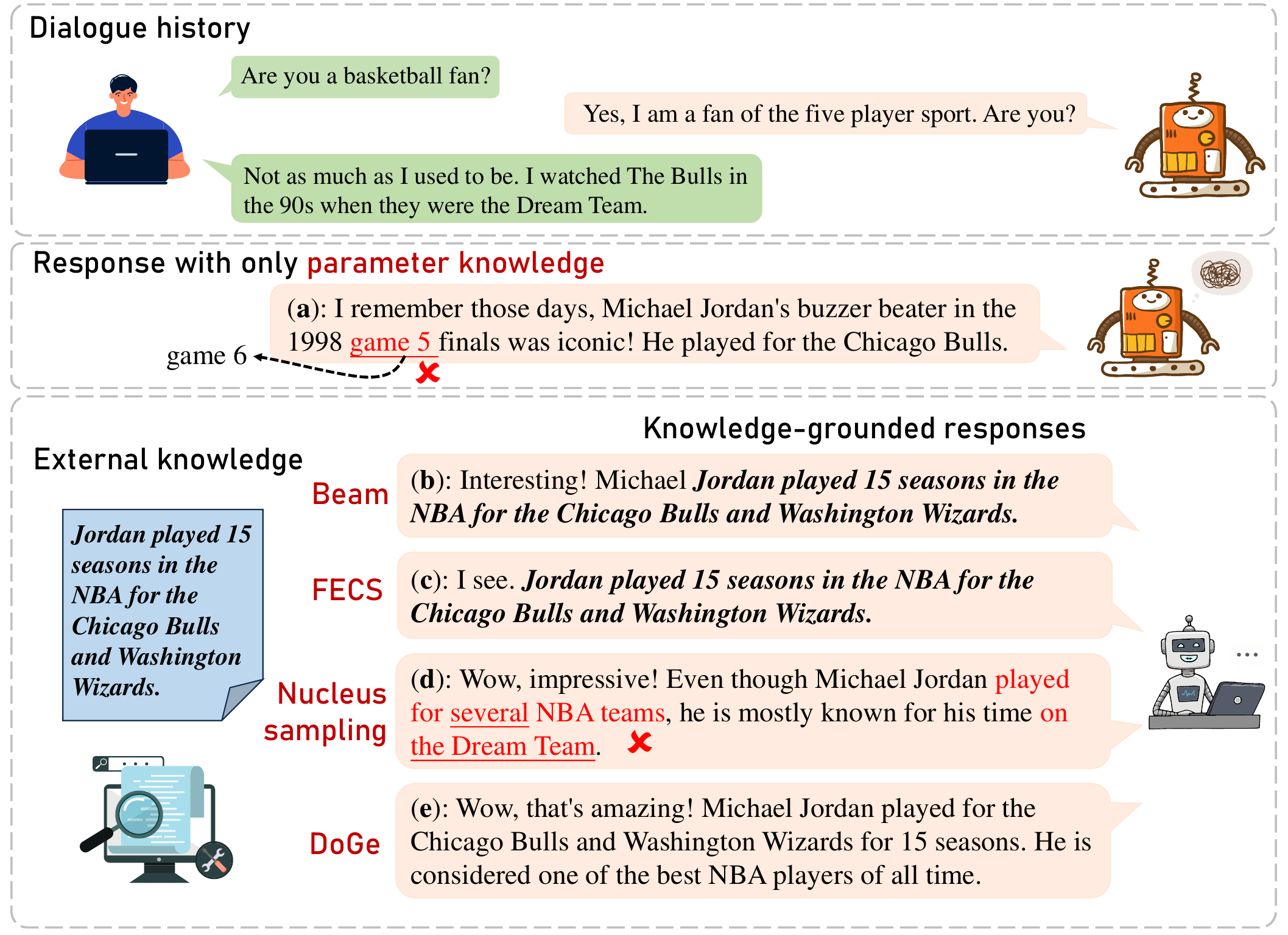}}
  \caption{An example illustrates the knowledge copying issue that may arise from increasing faithfulness to external knowledge and the hallucination drawback of sampling decoding remedy.}
  \label{fig-example}
\end{figure} 

Equipped with extensive parameters, large language models (LLMs) showcase robust generalization capabilities in various downstream NLP tasks \cite{touvron2023llama,workshop2023bloom,bai2023qwen,yang2023baichuan,touvron2023llama2}. Notably, their capacity in generating fluent and coherent content is comparable to that of humans in the domain of dialogue \cite{chatgpt,openai2023gpt4}. One conspicuous vulnerability of LLMs is the generation of non-factual responses \cite{Ji_2023}. Factuality is a crucial property in conversations, referring to the faithfulness of the generated content to the established world knowledge \cite{zhang2023sirens}. A plausible solution to enhance factuality is to provide the model with carefully selected external knowledge during response generation. Such an approach is in line with the Knowledge-Grounded Dialogue Generation (KGDG) task, for which several datasets have been introduced \cite{WoW,Holl-E,dziri-etal-2022-faithdial}. Furthermore, some studies have devised schemes to ensure the generated contents are more faithful to the provided knowledge \cite{deng-etal-2023-towards,sun2022contrastive,chen-etal-2023-fidelity}.



However, enhancing factuality through the faithfulness-augmented methods leads to the following drawbacks: (1) Excessive focus on the singular golden knowledge can result in the reduction of content diversity; (2) The overemphasis on faithfulness may diminish the engaging phrasing of responses, manifested as the model takes the shortcut of inserting portions of source knowledge into responses, such as the response (b) and (c) in Figure \ref{fig-example} \cite{yang-etal-2023-multi-level,daheim2023elastic}. Remedial actions could be adjusting the decoding strategy, as conventional deterministic decoding methods are prone to getting trapped in such text degeneration issues \cite{ul}. Current methods aiming at increasing diversity typically depend on the introduction of randomness in sampling \cite{holtzman2020curious}. Unfortunately, these stochastic methods are inclined to produce non-factual responses due to the selection of low-probability tokens \cite{lee2023factuality}. \textbf{It appears that there is a contradiction between factuality and diversity, and it is difficult to achieve the two goals simultaneously}. We empirically analyze the phenomenon and endeavor to reconcile the two more delicately, finding a way to promote diversity without compromising factuality.

In this paper, we propose a novel \textbf{D}ynamic s\textbf{o}urce-\textbf{G}rounded d\textbf{e}coding (DoGe) method. The fundamental idea is to reconcile the utilization of internal parameter knowledge and external source knowledge during generation, to enhance response diversity while maintaining the factuality. Existing decoding strategies are start-to-finish and incapable of addressing the dynamic demand for external knowledge of the model during generation. DoGe takes the factual confidence of the model as the foundation and dynamically switches the decoding strategy, thereby organically combining the sampling decoding and faithfulness-augmented deterministic decoding strategies. Specifically, DoGe dynamically switches between masking and exposing external knowledge to obtain two probability distributions based on the model's factual confidence. The local confidence and global uncertainty are integrated to measure the model's factual confidence, which acts as a proxy for the factuality of the generated content. If the LLM exhibits high factual confidence in its predictions without resorting to external knowledge, we deem its output factually accurate. Meanwhile, the masking is sustained and generation solely depends on parameter knowledge to increase diversity. On the contrary, if the LLM shows low confidence, DoGe exposes the external knowledge within the input and resorts to it for the prediction. Additionally, we design a scorer to re-rank candidate tokens by considering knowledge-attentive rewards during generation. This approach retains tokens that are sufficiently faithful to the knowledge source, thereby ensuring the content's faithfulness without fabrication.

Our contributions are summarized as follows:
 \begin{itemize}
	\item We probe the trade-off between faithfulness and diversity in current dialogue generation methods. Simply enhancing faithfulness results in damage to diversity, while compensating diversity through stochastic decoding causes damage to factuality.
	\item We propose an innovative \textbf{D}ynamic s\textbf{o}urce-\textbf{G}rounded d\textbf{e}coding (DoGe) method, which effectively reconciles the diversity and factuality in Knowledge-Grounded Dialogue Generation.
	\item We conduct comprehensive experiments, and the experimental results demonstrate the superiority of our method in both automated and human evaluation metrics.
\end{itemize}

\section{Related Work}
\subsection{Knowledge-grounded dialogue generation}
Knowledge-grounded dialogue generation aims to alleviate dull and unfaithful responses by infusing external knowledge into the input of dialogue models, and it encompasses two sub-tasks: knowledge selection and response generation. The hotspot of early research was primarily focused on how to enhance the performance of knowledge selection \cite{eacl-2023-generative,xu-etal-2022-corefdiffs, 2021-colv,yang-etal-2022-take,MIKE,kim2020SKT,wang-etal-2025-sibyl,yang-etal-2023-multi-level}. With the substantial leap in the capabilities of generative models, the research focus gradually shifts to the response generation sub-task \cite{zhao2020lowresource,liu2021threestage,zhao2020knowledgegrounded,zheng-etal-2021-knowledge,chen2024multitaskroleplayingagentcapable}. Ideally, an outstanding chat-bot should generate informative and truthful responses while maintaining the naturalistic phrasing and excellent interactivity \cite{dziri-etal-2022-faithdial}.
\par
\subsection{Hallucinations in Text Generation}

Hallucination pertains to the issue wherein the generated contents deviate from the user's instruction, contradict the previously generated context, or contravene the established world \cite{zhang2023sirens}. It presents severe risks to the practical applications in NLP. Existing works have conducted comprehensive explorations on hallucination from multiple viewpoints, including its origin \cite{mckenna2023sources, dziri2022origin}, detection \cite{zhang2023enhancing, manakul2023selfcheckgpt, fadeeva2023lmpolygraph}, and mitigation \cite{choi2023kcts, chuang2023dola, li2023inferencetime,yang2024orthogonalfinetuningdirectpreference}. Numerous approaches and benchmarks regarding hallucination were proposed, such as question answering \cite{gao-etal-2023-rarr, lin-etal-2022-truthfulqa}, text summarization \cite{cao-etal-2020-factual, zhong-etal-2021-qmsum}, and dialogue generation \cite{li2023halueval, wan2023sequencelevel, chen-etal-2023-fidelity, sun2022contrastive,Wang2022seek}. One effective solution to alleviate hallucination in QA is to rely on external knowledge as supplementary evidence, which provides more precise answers to user inquiries. Given the nature of open-domain conversations, a chit-chat agent should integrate external knowledge seamlessly into its responses, thereby attaining a balance between factuality and content diversity.

\section{Preliminaries}

\subsection{Task Formulation}

Given a task-specific instruction $\mathcal{I}$ of knowledge-grounded dialogue generation, a dialogue history $h$ across previous dialogue turns, a current user utterance $u$, and related knowledge sentence $k=(k_1,\dots,k_{m})$, where $m$ represents the number of tokens in $k$, the objective of the KGDG task is to generate an informative response $y=(y_1,\dots,y_n)$, with $n$ being the number of tokens in $y$. 
The conversation context $x$ is constructed by concatenating the task-specific instruction $\mathcal{I}$, the dialogue history $h$, and the user's last utterance $u$:
\begin{equation}
x = [\mathcal{I}; h; u].
\end{equation}
The probability distribution of the current token is derived by inputting the conversation context, corresponding knowledge, and generated tokens into a large language model:
\begin{equation}
p(y_t|x, k, y_{\leq t-1})=\mathcal{LLM}(x, k, y_{\leq t-1}).
\end{equation}
Ultimately, the response $y$ is generated via either a deterministic or stochastic decoding strategy.

\subsection{Faithfulness and Factuality}


Existing approaches alleviate the generation of non-factual responses by enhancing their faithfulness to external knowledge, which is overly restrictive and leads to the sacrifice of diversity \cite{honovich-etal-2021-q2, deng-etal-2023-towards}. In fact, it is merely one means of improving factuality. Models should be permitted to generate using factually correct parametric knowledge which is distinct from external knowledge. Faithfulness refers to the response being consistent with descriptions from retrieved external knowledge, with no conflicts, while factuality refers to the response being consistent with descriptions from world knowledge. We formally define faithfulness and factuality and explain their relationship to facilitate the research:

\noindent\textbf{Definition 3.1} Faithfulness($\mathcal{F}$):
\textit{Given a response} $y$, \textit{and external knowledge} $\mathcal{K} = (k_1, \ldots, k_j)$ \textit{at turn} $n$, \textit{we say that} \textit{the response $y$ is faithful with respect to the external knowledge} ($\mathcal{F}(\mathcal{K},y)$) \textit{if and only if the following condition holds}:
\begin{itemize}
    \item  $\exists\, \Gamma$ \textit{such that} $\Gamma \models y$, \textit{where} $\Gamma$ \textit{is a non-empty subset of} $\mathcal{K}$ \textit{and} $\models$ \textit{denotes semantic entailment}. \textit{In other words, there is no interpretation} $\mathcal{I}$ \textit{such that all members of} $\Gamma$ \textit{are true and} $y$ \textit{is false} \cite{dziri-etal-2022-faithdial}.
\end{itemize}
\noindent\textbf{Definition 3.2} Factuality($\mathcal{F}^*$):
\textit{Given a response} $y$, \textit{we say that} $y$ \textit{is factual} ($\mathcal{F}^*(y)$) \textit{if and only if the following condition holds}:
\begin{itemize}
    \item  $\exists\, \Phi$ \textit{such that} $\Phi \models y$, \textit{where}  $\Phi$ \textit{is a non-empty subset of world knowledge} $\mathcal{K}_w$ \textit{and} $\models$ \textit{denotes semantic entailment}.
\end{itemize}
\noindent\textbf{Theorem 3.1} \label{theorem} $\mathcal{F} \models \mathcal{F}^*, \mathcal{F}^* \not\models \mathcal{F}$, where $\models$ denotes entailment.


Theorem 3.1 indicates that responses ensuring faithfulness are necessarily factual, but the converse does not always hold. It also indicates that a method guaranteeing faithfulness can serve as a fallback for factuality, which provides some inspiration for DoGe. The proof of the theorem is presented in the Appendix A.

\subsection{Faithfulness-Diversity Trade-Off}


Given that such text degeneration issues could be caused by conventional deterministic decoding methods, to circumvent them, we explore the commonly used sampling-based methods to enhance the diversity and creativity of the generated content. We endeavor to balance faithfulness and diversity by controlling the degree of randomness. Specifically, we achieve that through adjustments of the temperature parameter $t$ to control the smoothness of the distribution and the threshold $p$ used in nucleus sampling to filter out low-probability token tails.

\begin{figure}[htbp]
  \centerline{\includegraphics[scale=0.29]{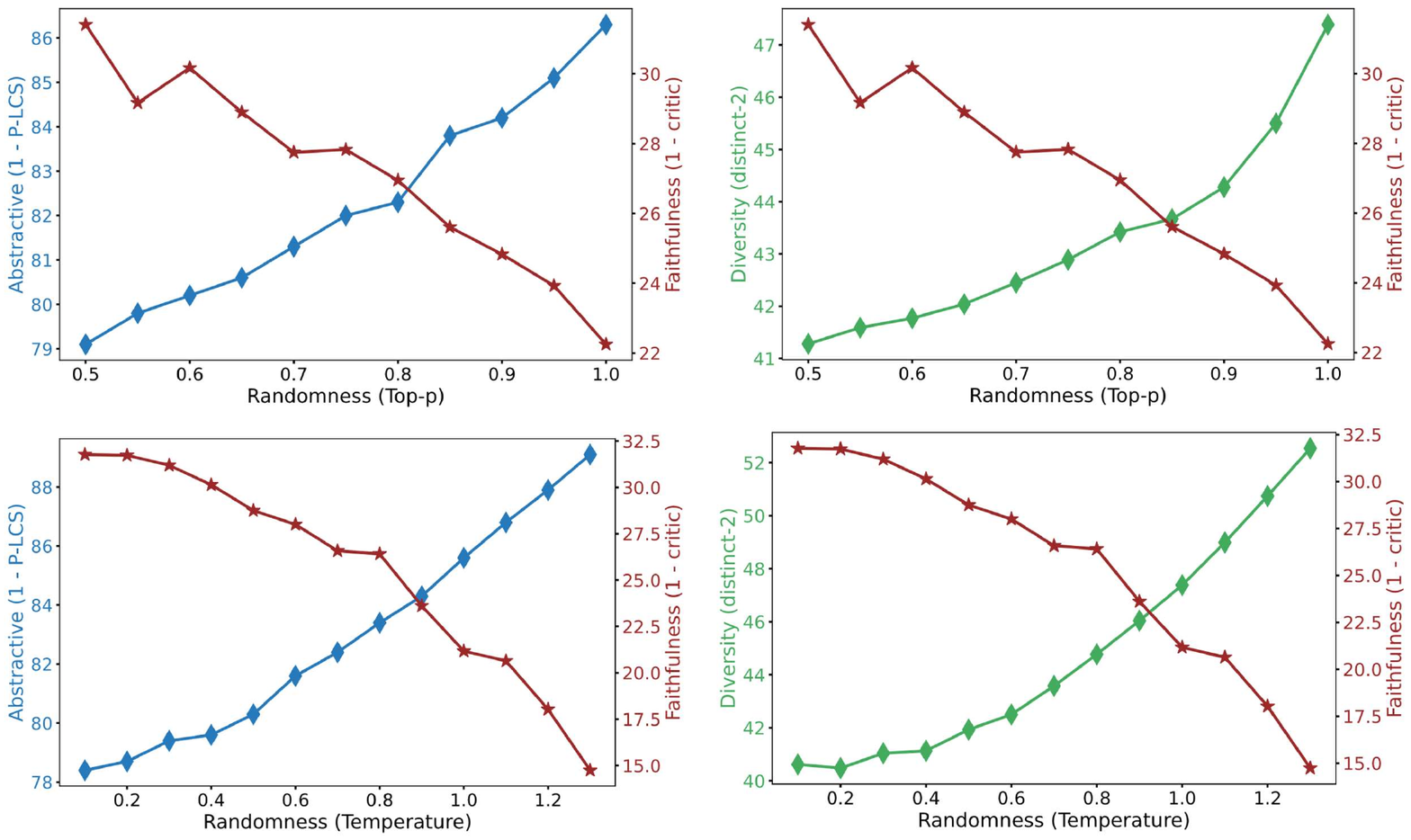}}
  \caption{Trade-off between faithfulness and diversity in different settings on the WoW(seen) dataset. }
  \label{figpilot}
\end{figure} 


We approximate faithfulness using the \textbf{Critic} metric proposed by \citet{dziri-etal-2022-faithdial}, assess content diversity through \textbf{distinct-2} and Precision of the Longest Common Sub-sequence (\textbf{P-LCS}) \cite{yang-etal-2023-multi-level}, where lower distinct-2 values indicate reduced diversity and higher P-LCS values indicate lower creativity of knowledge utilization. The experimental results, as depicted in Figure \ref{figpilot}, reveal an apparent trade-off between the two aspects. The trends show that a pursuit of enhanced diversity and creativity inevitably leads to an obvious reduction in faithfulness, which is inherently caused by the characteristic of stochastic decoding. \textbf{Our objective is to explore a way beyond the introduction of randomness to improve diversity without factual compromise}.

\section{Approach}

\begin{figure*}[htbp]
  \centerline{\includegraphics[scale=0.33]{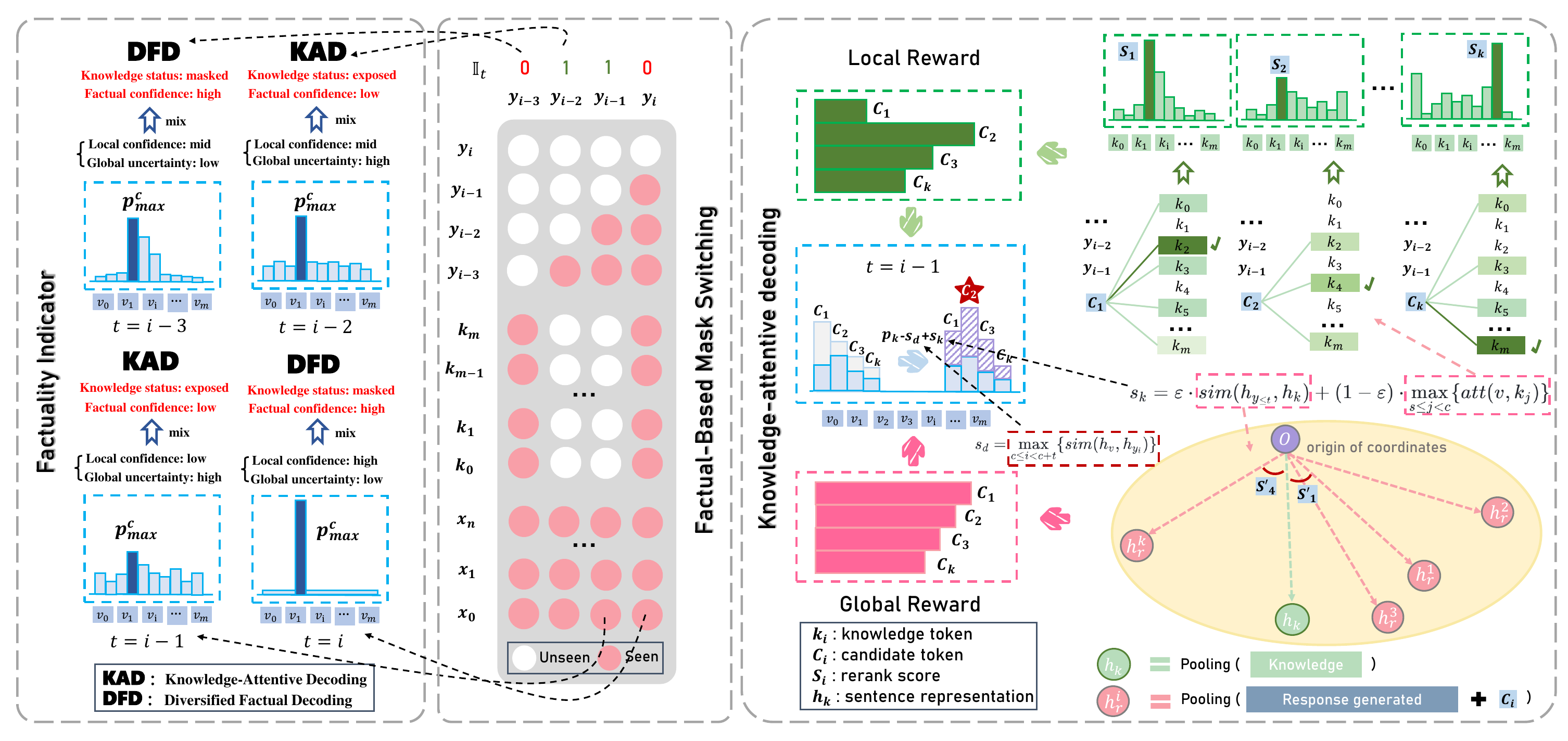}}
  \caption{An overview of the DoGe method. The left part of the left figure presents the results of several possible distributions of factual confidence scores and their corresponding decoding methods. The attention map in the right part of the left figure demonstrates how DoGe performs mask switching. The right part of the figure depicts the re-ranking process of KAD.}
  \label{figmethod}
\end{figure*}

\subsection{DoGe Overview}

The entire workflow of DoGe can be outlined as follows: 
DoGe initially computes a factual confidence score for the generated token based on the factuality indicator. Then, corresponding decoding methods are implemented at different factual confidence scores to achieve a balance of diversity and factuality. When the confidence score is high, we adopt the Diversified Factual Decoding strategy with external knowledge masked, granting the model higher freedom and fully mobilizing its parameter knowledge for more diversified expression. On the contrary, if the confidence score is low, DoGe resorts to the external knowledge within the input to recalculate the probability distribution for the prediction and employs Knowledge-Attentive Decoding to ensure the faithfulness of the generation. The overview of DoGe is illustrated in Figure \ref{figmethod}.

\subsection{Factual-Based Mask Switching} \label{subsection3.1}


Given that the incorporation of external knowledge might lessen the model's attention on the context and foster the undesirable act of knowledge copying \cite{yang-etal-2023-multi-level}, we adopt a dynamic mask-switching strategy on the external knowledge to disrupt the shortcut of copying and thereby promote creative knowledge integration. This strategy entails the dynamic alternation between masking and exposing external knowledge, guided by a factuality indicator. Through this approach, DoGe attains a balance in the utilization of parameter knowledge and external knowledge during generation. During generation, DoGe computes the two distributions $p(y_t|x, y_{\leq t-1})$ and $p(y_t|x, k, y_{\leq t-1})$ in parallel and dynamically switches the decoding strategy based on the factual confidence of the model. For the sake of writing convenience, we concurrently list the simplified symbolic representations of the two probability distributions as follows:
\begin{equation}
\begin{aligned}
p_c(y_t)&=p(y_t|x, y_{\leq t-1}), \\
p_k(y_t)&=p(y_t|x, k, y_{\leq t-1}).
\label{eq: 1}
\end{aligned}
\end{equation}

\subsection{Factuality Indicator}


Uncertainty can act as a metric for ascertaining when to trust LLMs. A number of studies probed into the possibility of utilizing uncertainty as a means to detect hallucinations \cite{manakul2023selfcheckgpt,huang2023look,duan2023shifting}. Given that a sufficiently advanced LLM is expected to assign low probabilities to tokens that are prone to introduce hallucinated content, we devised a Factuality Indicator through a logits-based uncertainty approach.


Inspired by \cite{zhang-etal-2023-enhancing-uncertainty}, we contend that factual confidence ought to be contemplated from two standpoints: local confidence and global uncertainty. The local confidence is delineated as the highest probability among the candidate tokens, and the global uncertainty is defined as the entropy of the entire probability distribution.

\begin{equation}
\begin{aligned}
p_{max}=\max_{y_t\in\mathcal{V}}p(y_t).
\label{eq: 7}
\end{aligned}
\end{equation}

\begin{equation}
\begin{aligned}
\mathcal{H}_t={-\sum_{y_t\in\mathcal{V}}p(y_t)*\log_2(p(y_t))}.
\label{eq: 77}
\end{aligned}
\end{equation}

Next, it is necessary for us to identify a suitable function $f$ for synthesizing both the local confidence $p_{max}$ and the global uncertainty $\mathcal{H}_t$, thereby deriving the factual confidence score $\mathcal F_t$.
\begin{equation}
\mathcal F_{t}= f ( p_{max}, {\mathcal{H}_t }, \gamma).
\label{eq: 5}
\end{equation}

We commence by conducting a straightforward transformation of global uncertainty $\mathcal{H}_t$ to obtain global factual confidence $\frac{1}{\eta\cdot\mathcal{H}_t + 1}$, which confines the domain of the functional results within the range of 0 to 1. Subsequently, we explored several synthetic approaches, such as arithmetic mean, harmonic mean, and geometric mean. The most optimal results were attained via the geometric mean $\mathcal F^*_{t}$, while the other methods yielded marginally inferior performance. The definitions of the remaining synthetic methods are presented in the Appendix B.

\begin{table*}[]
\centering
\renewcommand\arraystretch{1.2}
\scalebox{0.65}{
\begin{tabular}{l|cccccccc|cccccccc}
\toprule
\multirow{2}{*}{\textbf{Method}} & \multicolumn{7}{c}{\textbf{Wizard of Wikipedia (seen)}}                                                                              &      & \multicolumn{7}{c}{\textbf{Wizard of Wikipedia (unseen)}}                                                                          &               \\ \cline{2-17} 
                                 & \textbf{BLEU-1/2}  & \textbf{MET} & \textbf{Dist-1/2}   & \textbf{Ent-1/2}            & \textbf{BS} & \textbf{Critic}         & \textbf{Fact.}         & \textbf{CFD}  & \textbf{BLEU-1/2}  & \textbf{MET} & \textbf{Dist-1/2} & \textbf{Ent-1/2}            & \textbf{BS} & \textbf{Critic}         &  \textbf{Fact.}         & \textbf{CFD}           \\ \hline
Greedy                           
& 27.1/16.3          & 15.6            & 7.63/40.30           & 6.42/9.48          & 55.5             & 31.51          & 87.0          & 35.63 
& 27.6/17.0            & 16.3            & 5.67/29.03        & 6.22/9.06          & 56.3             & 30.28          & 86.5          & 29.65          \\
Beam                            
& 29.1/18.9          & 16.7            & 8.29/41.89          & 6.53/9.53          & 59.2             & 42.76          & \textbf{87.5} & 42.32 
& 29.7/19.5 & 17.5            & 6.02/29.05        & 6.31/9.05          & 60.6             & \textbf{43.72} & {86.0} & 35.64 \\
Nucleus                         
& 24.9/14.1          & 14.5            & 7.77/44.61          & 6.51/9.77          & 52.5             & 25.27          & 81.5          & 33.58 
& 25.1/14.3          & 14.9            & 6.13/35.93        & 6.34/\textbf{9.47}          & 52.7             & 24.90           & 80.5          & 29.91          \\
F-Nucleus                       
& 25.6/14.7          & 14.7            & 7.74/43.07          & 6.47/9.66          & 53.2             & 26.36          & 84.0          & 33.69 
& 26.0/15.2            & 15.3            & 5.98/34.02        & 6.3/9.36           & 54.0               & 28.64          & 84.5          & 31.21          \\
DoLa                            
& 27.5/16.5          & 15.5            & 7.73/40.32          & 6.41/9.45          & 55.4             & 31.48          & 86.5          & 35.63 
& 28.0/17.2            & 16.2            & 5.71/29.02        & 6.21/9.03          & 56.2             & 30.37          & 86.0          & 29.69          \\
CD                               
& 23.2/13.0            & 13.9            & 9.23/\textbf{52.62}          & 6.64/9.91          & 52.3             & 26.44          & 79.5          & 37.30 & 23.8/14.2          & 15.2            & 6.79/\textbf{37.86}        & 6.43/9.45          & 53.1             & 27.53          & 81.5          & 32.28          \\
CS                               
& 26.2/15.3          & 14.8            & 8.70/44.44           & 6.58/9.68          & 54.0               & 31.12          & 86.0          & 37.19 
& 26.7/16.0            & 15.5            & 6.47/32.74        & 6.38/9.28          & 54.9             & 32.21          & 85.0          & 32.47          \\
FECS                             
& 29.3/18.4          & 16.5            & 8.47/43.28          & 6.54/9.62          & 59.5             & 35.94          & 86.5          & 39.44 
& 30.0/19.1            & 17.2            & 6.05/29.90         & 6.30/9.12           & 60.4             & 34.89          & 84.0          & 32.30          \\
\textbf{DoGe}               
& \textbf{31.1*/19.8*} & \textbf{17.6*}   & \textbf{9.77*}/49.58 & \textbf{6.79*/9.96} & \textbf{64.7*}    & \textbf{43.93*} & \textbf{87.5} & \textbf{46.67*}
& \textbf{31.2*/20.1*} & \textbf{18.0*}     & \textbf{7.24*}/34.67        & \textbf{6.54*}/9.42 & \textbf{65.6*}    & 43.21          & \textbf{87.0} & \textbf{38.71*}          \\ \bottomrule
\end{tabular}
}
\caption{Automatic Evaluation results on the WoW dataset (LLaMA2-chat-7B). 
The best results are highlighted with \textbf{bold}. "*" denotes that the improvement to the best baseline is statistically significant (t-test with $p$-value $<$  0.01). }
\label{tab: auto1}
\end{table*}

\begin{table*}[]
\centering
\renewcommand\arraystretch{1.2}
\scalebox{0.65}{
\begin{tabular}{l|cccccccc|cccccccc}
\toprule
\multirow{2}{*}{\textbf{Method}} & \multicolumn{7}{c}{\textsc{FaithDial}}                                   &      & \multicolumn{7}{c}{\textbf{Holl-E}}                                                                 &               \\ \cline{2-17} 
                                & \textbf{BLEU-1/2}  & \textbf{MET} & \textbf{Dist-1/2}   & \textbf{Ent-1/2}            & \textbf{BS} & \textbf{Critic}         & \textbf{Fact.}         & \textbf{CFD}  & \textbf{BLEU-1/2}  & \textbf{MET} & \textbf{Dist-1/2} & \textbf{Ent-1/2}            & \textbf{BS} & \textbf{Critic}         &  \textbf{Fact.}         & \textbf{CFD}         \\ \hline
Greedy                           
& 28.6/17.2          & 17.8            & 7.97/38.96          & 6.40/9.34           & 58.9             & 36.52          & 89.0          & 37.73 
& 38.2/29.3          & 20.5            & 5.49/28.46        & 5.95/8.80  & 55.5             & 32.59          & \textbf{95.5}          & 30.46          \\
Beam                             
& 31.0/19.5            & 18.8            & 8.63/39.99          & 6.48/9.35          & 62.3             & 46.24          & {89.5} & 43.00 
& \textbf{44.3/36.4} & 24.4            & 6.16/30.54        & 6.16/8.99 & 59.4             & \textbf{39.93} & {95.0} & {34.92} \\
Nucleus                          
& 26.0/14.8            & 16.4            & 8.23/43.54          & 6.47/9.61          & 54.9             & 29.15          & 79.5          & 35.63 
& 32.1/22.4          & 17.5            & 5.84/34.24        & 6.14/9.27 & 51.9             & 18.78          & 91.0          & 25.36          \\
F-Nucleus                        
& 26.9/15.5          & 16.7            & 8.30/42.30            & 6.44/9.53          & 55.9             & 31.25          & 83.0          & 36.36 
& 33.7/24.5          & 18.5            & 5.76/32.72        & 6.10/9.15  & 53.1             & 20.86          & 92.5          & 26.13          \\
DoLa                             
& 29.0/17.5            & 17.7            & 8.03/38.89          & 6.38/9.31          & 58.7             & 36.34          & 87.0          & 37.59 
& 38.3/29.3          & 20.3            & 5.50/28.42         & 5.95/8.78 & 55.5             & 32.54          & \textbf{95.5}          & 30.41          \\
CD                               
& 25.4/14.5          & 15.9            & 9.67/\textbf{51.33}          & 6.61/\textbf{9.76}          & 55.1             & 32.74          & 78.0          &  40.99 
& 33.9/26.1          & 17.7            & 6.25/\textbf{38.94}        & 6.11/9.22 & 51.6             & 19.04          & 89.5          & 27.23          \\
CS                               
& 27.9/16.3          & 16.8            & 9.11/43.17          & 6.53/9.53          & 56.9             & 33.49          & 84.5          & 38.02 
& 34.6/25.4          & 18.5            & 6.26/31.35        & 6.22/9.11 & 53.2             & 26.35          & 92.0          & 28.74          \\
FECS                             
& 30.7/18.9          & 18.6            & 8.61/40.80           & 6.48/9.41          & 62.4             & 41.01          & 88.0          & 40.90 
& 42.0/33.3            & 22.5            & 5.97/30.34        & 6.07/8.94 & 57.7             & 36.71          & 94.5          & 33.37          \\
\textbf{DoGe}              
& \textbf{32.7*/20.2*} & \textbf{19.9*}   & \textbf{10.17*}/47.50 & \textbf{6.74*/9.76} & \textbf{67.9*}    & \textbf{48.01*} & \textbf{91.0} & \textbf{47.75*} 
& 44.0/36.0              & \textbf{25.3*}   & \textbf{6.77*}/35.27        & \textbf{6.38*/9.37*} & \textbf{59.7}    & 37.43          & \textbf{95.5} & \textbf{36.33*}          \\ \bottomrule
\end{tabular}
}
\caption{Automatic Evaluation results on the \textsc{FaithDial} and {Holl-E} dataset (LLaMA2-chat-7B). 
The best results are highlighted with \textbf{bold}. "*" denotes that the improvement to the best baseline is statistically significant (t-test with $p$-value $<$ 0.01). }
\label{tab: auto2}
\end{table*}

\begin{equation}
\mathcal F^*_{t}= \sqrt[2]{\frac{ p_{max}}{\eta\cdot\mathcal{H}_t + 1}  }- \gamma,
\label{eq: m0}
\end{equation}
where $\eta$ is a hyper-parameter that governs the strength of $\mathcal{H}_t$ (set to 1) and $\gamma$ is a pre-defined threshold representing reliable factuality.

Firstly, we compare the factual confidence scores of the model in the circumstances of masking $F^c_{t}$ and exposing external knowledge $F^k_{t}$. If the score declines upon provision of knowledge, it implies that the external knowledge is not relevant to the generation, or the model generates non-knowledgeable content at the current step. When the provision of external knowledge enhances the factual confidence score and if the score exceeds the factual threshold, it indicates that the parameter knowledge of LLM is sufficient to handle the user's query. Otherwise, DoGe employs knowledge-attentive decoding to enhance the faithfulness of generation.
\begin{equation}
 \mathbb I_{t}=\left\{ \begin{array}{ll}
1 & \mbox{if} \  (\mathcal F^c_{t} > 0) \; \vee \; ( F^k_{t} -  F^c_{t} < 0 ),  \\[1ex]
0 & \mbox{otherwise}.  \\ 
\end{array} \right.
\label{eq: 4}
\end{equation}


Since $F^c_{t}$ and $ F^k_{t}$ are derived from $p_c(y_t)$ and $p_c(y_t)$ that will be reused subsequently, the design of DoGe's Factuality Indicator incurs negligible additional overhead. Additionally, we can regulate DoGe's preference for the diversity or factuality of the generated content by adjusting hyper-parameters in diverse generation scenarios.

\subsection{Diversified Factual Decoding} \label{dfd}

When the factual confidence score is high, we employ the Diversified Factual Decoding strategy with external knowledge masked, granting the model greater freedom to fully mobilize its parameter knowledge for more diversified expressions. The Diversified Factual Decoding strategy is implemented through top-P sampling, which filters out unreliable candidate tokens with low probability.
\begin{equation}
\sum_{y_t\in\mathcal{V}^{(p)}}p_c(y_t) \geq \tilde{P}.
\label{eq: 2}
\end{equation}

\begin{equation}
  \hat{p}_c(y_t)=\left\{ \begin{array}{ll}
p_c(y_t)/\hat{P} & \mbox{if} \  y_t \in \mathcal{V}^{(p)}  \\[1ex]
0 & \mbox{otherwise},  \\ 
\end{array} \right.
\label{eq: 24}
\end{equation}
where $\hat{P} = \sum_{y_t\in\mathcal{V}^{(p)}}p_c(y_t)$. Finally, we obtain $y_t^c$ by sampling with the revised probability distribution $\hat{p}_c(y_t)$.

\subsection{Knowledge-Attentive Decoding} \label{subsection3.2}

When Doge demonstrates a deficiency in factual confidence, it resorts to external knowledge. We designed a scoring system to re-rank the top-K candidate tokens, giving preference to those that are more faithful to the external knowledge and unique. The re-ranking process of the top k candidates is carried out in accordance with the following formula:
\begin{equation}
\hat{p}_k(y_t)=(1-\alpha-\beta) \cdot p_k(y_t) - \alpha  \cdot s_d + \beta \cdot s_k,
\label{eq: 255}
\end{equation}
where $y_t \in \mathcal{V}^{(K)} $, $s_d$ indicates degeneration inhibition score and $s_k$ indicates knowledge attentive score.


We guarantee faithfulness in two respects: (1) Token-level reward: The scoring system encourages selecting tokens that pay greater attention to the knowledge segment. (2) Sentence-level reward: The scoring system prefers tokens that render the entire response more semantically coherent with external knowledge. For the sentence-level reward, we concatenate the candidate token with the previously generated tokens to form candidate sentences. Subsequently, the sentence representations of the knowledge and responses are computed by mean-pooling their hidden states. The cosine similarity between these sentence representations is regarded as the reward. Considering that responses in the initial stage are incomplete and the sentence-level reward is of less significance, we set a weighting parameter $\varepsilon$ to balance the two rewards. The influence of sentence-level reward scores is associated with the length of the generated content. The entire design of the knowledge attentive score is presented as follows:
\begin{equation}
s_k = \varepsilon \cdot sim(h_{y_{\leq{t}}},h_k)  + (1-\varepsilon) \cdot \max\limits_{s\leq j < c}  \{att(v,{k_j})\},
\end{equation}
where $sim(\cdot,\cdot)$ denotes the cosine similarity, $att(v,k_j)$ denotes the attention weight between $v$ and $k_j$ after max-pooling for all the layers and attention heads, and $\varepsilon$ is a hyper-parameter balancing two rewards.
\begin{equation}
\varepsilon = \max \{ \omega,\lambda^{\frac{t}{N}-1}\},
\end{equation}
where $\lambda$ is a growth factor, $\omega$ controls the max growth rate, and $N$ denotes the max generation length.

To prevent dull and repetitive degeneration, we adopt the penalty term from contrastive search which inhibits repetition of what has been generated \cite{su2022contrastive}.
\begin{equation}
s_d =  \max\limits_{c\leq i < c+t}  \{sim(h_v,h_{y_i})\}.
\end{equation}
The predicted token $y_t^k$ is finally obtained by greedy search:
\begin{equation}
y_t^k = \arg\max\limits_{y_t \in \mathcal{V}^{(K)}} \hat{p}_k(y_t).
\end{equation}

\subsection{Final Prediction} \label{subsection3.3}

We combine the aforementioned designs to obtain the ultimate distribution by employing the formula below:
\begin{equation}
y_t=\mathbb I_{t} \cdot y_t^c + (1-\mathbb I_{t}) \cdot y_t^k.
\label{eq: 8}
\end{equation}
Refer to Appendix for the pseudo-code of the entire decoding process.

\section{Experiments}

\subsection{Experimental Setup}

\textbf{Dataset.}
We conduct experiments on three knowledge-grounded dialogue datasets: Wizard of Wikipedia (WoW) \cite{WoW}, \textsc{FaithDial} \cite{dziri-etal-2022-faithdial}, and Holl-E \cite{Holl-E}. \textbf{WoW} is collected on the basis of Wikipedia, where one crowd-sourcer assumes the role of a knowledgeable wizard while the other takes on the part of an inquisitive apprentice. We evaluate it on both its test \texttt{seen} set and test \texttt{unseen} set. 
\textsc{FaithDial} is a novel benchmark for hallucination-free dialogues, which optimizes the responses in the WoW dataset to be more faithful to knowledge. 
\textbf{Holl-E} is constructed by MTurk workers and focuses on movies as two workers engage in conversations about a chosen movie with each other. The quality of the last dataset is poorer than that of the previous two, as there are numerous samples with highly similar ground-truth responses and knowledge. We primarily focus on the evaluation results of the first two datasets.

\noindent \textbf{Baselines.}
We compare DoGe with the following decoding strategy baselines: Greedy Decoding (Greedy), Beam Search (Beam), Nucleus Sampling (Nulceus), Factual-Nucleus Sampling (F-Nucleus), DoLa, Contrastive decoding (CD), Contrastive search (CS), and FECS. The detailed introductions of these baselines are given in the Appendix C.

\noindent \textbf{Implementation Details.} We selected the prevalent Llama-2-7b-chat-hf and Llama-2-13b-chat-hf model \cite{touvron2023llama2} as the backbone model and carried out experiments on them, making use of the open-source Hugging Face transformers \cite{transformers}. All experiments were conducted with few-shot prompting (three shots). The three demonstrations were randomly picked from the dataset, and they are presented along with task instructions in the Appendix E. We omitted the step of knowledge selection and directly utilized manually annotated golden knowledge from the three datasets as input in the experiments. For the hyper-parameters in Doge, we set $\alpha|\beta|\lambda|\omega|K|\tilde{P}=0.4|0.35|0.8|0.4|4|0.9$.


\subsection{Experimental Results}

\subsubsection{Automatic Evaluation}


We chose the \textbf{BLEU} \cite{2002-bleu} and \textbf{METEOR} \cite{banerjee-lavie-2005-meteor} for evaluating the generation quality, Distinct-n (\textbf{Dist-n}) \cite{li-etal-2016-diversity} and Entropy (\textbf{Ent-n}) \cite{zhang2018generatinginformativediverseconversational} for assessing the generation diversity, BertScore (\textbf{BS}) \cite{zhang2020bertscore} and \textbf{Critic} \cite{dziri-etal-2022-faithdial} for measuring the faithfulness of the generated responses to external knowledge. Additionally, we randomly sampled 200 instances from the test set and evaluated their factuality by instructing GPT-4 in the manner of G-Eval \cite{G-EVAL}, denoted as \textbf{Fact.}. The \textbf{Critic} score reflects the proportion of samples that are faithful to external knowledge, while the \textbf{Fact.} value indicates the proportion of samples that are faithful to world knowledge among the 200 samples. We also devised a combination of faithfulness and diversity (\textbf{CFD}) metric that computes the harmonic mean of the \textbf{Critic} and \textbf{Dist-2} scores to manifest the ability of these methods in harmonizing the two evaluation criteria.

\begin{figure}[htbp]
  \centerline{\includegraphics[scale=0.32]{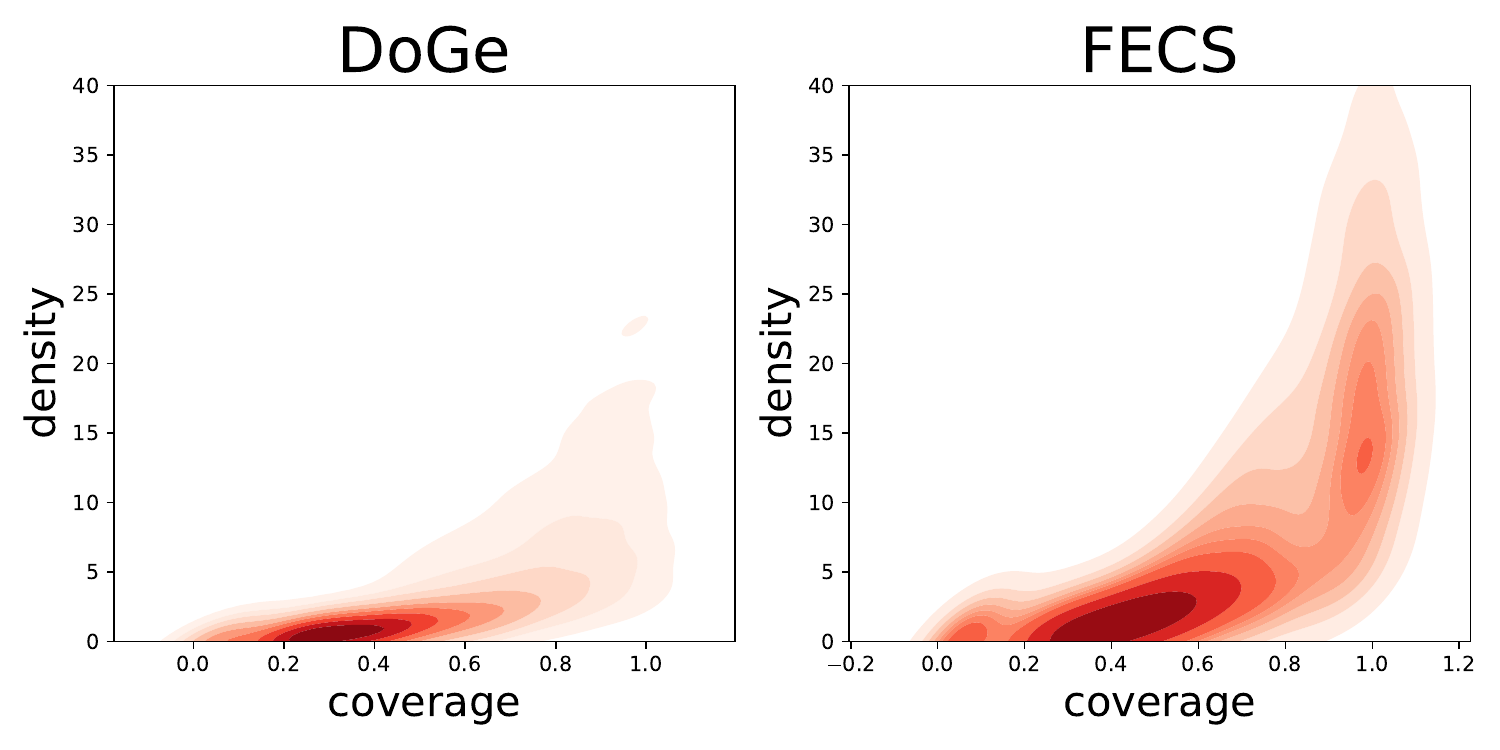}}
  \caption{Density and coverage comparison. DoGe's generation tends to be more abstractive than that of FECS.}
  \label{fig333}
\end{figure}

Tables \ref{tab: auto1} and \ref{tab: auto2} present the outcomes on the Llama-2-7b-chat model across three datasets. We will illustrate the efficacy of DoGe from three aspects. Firstly, DoGe attains the \textbf{SOTA} results on the \textbf{CFD} and \textbf{Fact.} metrics in all settings. It strikes an optimal balance between the utilization of internal and external knowledge, effectively enhancing diversity while maintaining the prerequisite of factuality. Secondly, DoGe achieves considerable improvements on both \textbf{BLEU-1/2} and \textbf{MET} metrics, outperforming all baselines. This enhancement indirectly demonstrates DoGe's effective reconciliation of diversity and factuality, which leads to higher consistency with the ground-truth across the overlap-based metrics evaluating generation quality. Lastly, the scaling experimental results in Tables \ref{tab: auto3} and \ref{tab: auto4} indicate that DoGe still functions well on larger-scale models. What is more exciting is that DoGe's performance on the 7b LLM is equal to or even exceeds some baseline's performance on the 13b LLM, which showcases its potential value in application scenarios. In addition to content diversity, the natural integration of knowledge is also a significant aspect of diversity. To evaluate it, we adopted the distribution between coverage and density proposed by \citet{grusky2020newsroom}, where a high density indicates a tendency to extract snippets from external knowledge for constructing responses. According to the results depicted in Figure \ref{fig333}, compared to the strongest baseline FECS, DoGe's application of knowledge is more creative, maintaining natural phrasing. These findings suggest that DoGe is a decoding strategy that achieves the best overall performance in open-domain dialogue.

\begin{table*}[]
\centering
\renewcommand\arraystretch{1.2}
\scalebox{0.67}{
\begin{tabular}{l|cccccccc|cccccccc}
\hline
\multirow{2}{*}{\textbf{Method}} & \multicolumn{7}{c}{\textbf{Wizard of Wikipedia(seen)}}                                                                             &                & \multicolumn{7}{c}{\textbf{\textsc{FaithDial}}}                                                                            &                \\ \cline{2-17} 
                                 & \textbf{BLEU-1/2}  & \textbf{METEOR} & \textbf{Dist-1/2}   & \textbf{Ent-1/2}   & \textbf{BS}   & \textbf{Critic} & \textbf{Fact.} & \textbf{CFD}   & \textbf{BLEU-1/2}  & \textbf{METEOR} & \textbf{Dist-1/2}    & \textbf{Ent-1/2}   & \textbf{BS}   & \textbf{Critic} & \textbf{Fact.} & \textbf{CFD}   \\ \hline
\textbf{DoGe}                    & 31.1/19.8          & 17.6            & \textbf{9.77/49.58} & \textbf{6.79/9.96} & 64.7          & \textbf{43.93}  & \textbf{87.5}  & \textbf{46.67} & {32.7/20.2} & {19.9}   & \textbf{10.17/47.50} & \textbf{6.74/9.76} & 67.9          & \textbf{48.01}  & \textbf{91.0}  & \textbf{47.75} \\
-$\varepsilon$                         & 30.7/19.3          & 17.2            & 9.65/49.30          & 6.74/9.89          & 64.0          & 42.98           & \textbf{87.5}  & 46.03          & 32.3/19.7          & 19.5            & 10.03/47.22          & 6.69/9.69          & 67.3          & 47.13           & \textbf{91.0}  & 47.17          \\
-$s_k(s)$                           & 30.9/19.5          & 17.3            & 9.62/49.14          & 6.72/9.84          & 64.2          & 40.56           & 87.0           & 44.64          & 32.3/19.6          & 19.6            & 10.07/47.06          & 6.67/9.64          & 67.3          & 44.54           & \textbf{91.0}  & 45.78          \\
-$s_k(t)$                           & 29.4/18.0          & 16.6            & 9.69/48.86          & 6.77/9.91          & 59.8          & 36.27           & 86.5           & 42.10          & 31.1/18.4          & 18.9            & 10.06/46.88          & 6.72/9.71          & 63.6          & 40.38           & 90.0           & 43.51          \\
-$s_d$                              & 31.4/19.9          & 17.7            & 8.79/45.81          & 6.54/9.80          & \textbf{65.3} & 42.64           & \textbf{87.5}  & 44.20          & 32.9/\textbf{20.4}          & \textbf{20.0}            & 9.18/43.71           & 6.49/9.6           & \textbf{68.5} & 46.42           & \textbf{91.0}  & 45.04          \\
-KAD                             & 26.8/16.2          & 15.8            & 8.51/44.97          & 6.53/9.76          & 54.0          & 30.95           & 86.0           & 37.31          & 28.8/16.9          & 18.1            & 8.91/42.89           & 6.48/9.56          & 57.1          & 35.07           & 89.0           & 38.78          \\
-FBM                             & \textbf{31.5/20.1} & \textbf{17.8}   & 9.02/45.43          & 6.58/9.68          & 65.1          & 43.72           & \textbf{87.5}  & 44.57          & \textbf{33.0}/20.2          & \textbf{20.0}            & 9.42/43.35           & 6.53/9.48          & 68.2          & 47.68           & 90.5           & 45.46          \\ \hline
\end{tabular}
}
\caption{Ablation study on the WoW (seen) and \textsc{FaithDial} dataset. -FBM indicates the elimination of the factual-based masking design and persistent generation based on KAD. -KAD implies the removal of all re-ranking scores in knowledge-attentive decoding. -$s_k(s)$ represents the elimination of the sentence-level reward. -$s_k(t)$ represents the removal of the token-level reward. -$\varepsilon$ indicates the elimination of the balancing hyper-parameter between two rewards and simply averaging them . -$s_d$ represents the elimination of the degeneration inhibition score. }

\label{tab:ablation}
\end{table*}
\subsubsection{LLMs-based Evaluation}
We utilize G-Eval to evaluate the Naturalness (\textbf{Nat.}) and Coherence (\textbf{Coh.}) of responses from five decoding methods, as elaborated in previous studies \citep{G-EVAL,closergeval}. Additionally, we formulate explicit instructions to assess the task-specific metric Informativeness (\textbf{Inf.}) by strictly adhering to the established rating strategy \cite{GPTscore}. The \textbf{Inf.} metric ascertains which response incorporates more intriguing knowledge. We prompted GPT-4 to assign discrete ratings ranging from 1 to 3 points to these generated responses. Table \ref{tab: G_eval} presents the outcomes of the LLMs-based evaluation, where the responses generated by DoGe outperform all baseline methods in terms of \textbf{Nat.}, \textbf{Inf.} and \textbf{Coh.}. This significant enhancement is attributed to DoGe's natural knowledge integration and an exceptional ability to balance the utilization of parameter knowledge with external knowledge.

\subsubsection{Human Evaluation}

For human evaluation, 100 samples were randomly selected from the test \texttt{seen} set of the WoW dataset. Given the context and its responses from the baseline models, five highly educated annotators were requested to select the superior response based on three criteria: (1) Coherence (\textbf{Coh.}): which model generates responses that are more contextually coherent; (2) Engagingness (\textbf{Eng.}): which model produces responses that are more interesting; and (3) Factuality (\textbf{Fac.}): which response contains fewer factual errors. To gauge the agreement among different annotators, the Fleiss’ kappa value \cite{Fleiss1971MeasuringNS} was computed. The results of the human-based evaluation are presented in Figure \ref{fighuman}. The findings reveal that our method surpasses the baselines in terms of factuality, while also guaranteeing that the generated content is both creative and diverse. Furthermore, the responses generated by DoGe demonstrate a strong relevance to the context, ensuring a smooth conversational flow.

\subsection{Analysis}

\textbf{Every design of DoGe is indispensable.} To evaluate the efficacy of DoGe, we carried out ablation studies by eliminating designs from it. The ablation results for the \textbf{WoW} and \textsc{FaithDial} datasets are presented in Table \ref{tab:ablation}. We discovered that the removal of any module leads to a deterioration in performance in various aspects. Specifically, the absence of KAD leads to a significant decline in every metric as it is a key design of DoGe. Removing either FBM or $s_d$ results in different levels of decrease in the diversity of the generated responses. Eliminating FBM suppresses the utilization of parameter knowledge, significantly reducing the creativity of the generated content, while eliminating $s_d$ leads to the frequent usage of common tokens. Furthermore, we find that both token-level and sentence-level rewards contribute positively to KAD, while the former has a slightly greater influence. The organic combination of both rewards guaranteed by $\varepsilon$ achieves more effective integration of external knowledge, while a casual combination could undermine the performance.

\noindent \textbf{Case study.} 
To intuitively evaluate the performance of diverse decoding approaches, we contrast the responses produced by Beam Search, FECS, Nucleus sampling, and DoGe. The instances presented in Figure \ref{fig-example} reveal that both Beam Search and FECS tend to solely replicate external knowledge, leading to a deficiency in engagement and a colloquial style. Owing to the nature of random sampling, Nucleus sampling introduces two incorrect facts: (1) Jordan actually played for only two teams, and (2) Jordan is mostly renowned for his tenure with the Bulls. In contrast, the response generated by DoGe not only showcases a higher degree of interactivity but also enriches the discussion by integrating information beyond the provided knowledge. Additionally, we implemented a simple yet intuitive visualization of knowledge usage in Figure \ref{fig-case}. The fact is inferred from external knowledge in the former sentence, while the fact in the last sentence is generated based on internal knowledge.

\noindent \textbf{Efficiency.} 
To appraise the decoding latency of DoGe, we present the average decoding time (sec) per instance on the \textsc{FaithDial} dataset in Table \ref{tabspeed}. The outcomes are averaged across all instances in the test set. Thanks to parallel implementation, DoGe operates marginally more slowly than beam search. Nevertheless, it generates faster than the re-rank-all-the-time approaches: CS and FECS.

\section{Conclusion}


In this paper, we uncover the finding that enhancing factuality by means of faithfulness-augmented approaches results in a decline in content diversity and the creative employment of knowledge, while remedial measures based on sampling compromise factuality. Consequently, we present a novel DoGe method, which strikes a balance between the utilization of parameter knowledge and external knowledge during generation. Extensive experiments validate that DoGe effectively mitigates the trade-off between factuality and diversity in knowledge-grounded dialogue generation.

\section*{Ethics Statement}

The benchmark datasets we employed in our experiments—WoW \cite{WoW}, \textsc{FaithDial} \cite{dziri-etal-2022-faithdial}, and Holl-E \cite{Holl-E}—are all highly regarded, open-source datasets amassed by crowd-sourced workers. They were assembled in strict compliance with user privacy protection protocols, guaranteeing the exclusion of any personal information. Furthermore, our proposed methodology is deliberately designed to uphold ethical norms and promote social fairness, ensuring that no bias is introduced. For the human evaluation aspect of our study, all participants were volunteers who were provided with comprehensive information regarding the purpose of the research, ensuring informed consent. Additionally, all participants were compensated fairly and reasonably for their contributions. We informed the annotators that the data is to be utilized solely for academic research purposes.

DoGe fails to validate the factual correctness of the offered reference, thereby entailing a risk of potential misuse. Specifically, in the event that the knowledge source encompasses non-factual information, DoGe is prone to generate misinformation as well. We earnestly recommend that users meticulously verify the knowledge source prior to its utilization. We adopt AI writing in our work, merely for refining articles to boost their readability.
\section{Acknowledgments}

\bibliography{aaai24}

\newpage
\appendix

\section{Appendix A  The proof of Theorem 2.1}
\label{proof}

\noindent\textbf{Theorem.} $\mathcal{F} \models \mathcal{F}^*, \mathcal{F}^* \not\models \mathcal{F}$.

\begin{proof}
$\mathcal{F} \models \mathcal{F}^*:$
For all $y$ that satisfy $\mathcal{F}(K,y)$, there exists $\Gamma$ such that $\Gamma \models y$ and $\Gamma \subseteq K$. Since $K \subsetneq K_{w}$ (external knowledge is a proper subset of world knowledge), it follows that $\Gamma \subseteq K_{w}$. Let $\Phi=\Gamma$, then  $\Phi \models y$ and $\Phi \subseteq K$. Hence, $\mathcal{F}^*(y)$ holds, and the conclusion is proved. 

$\mathcal{F}^* \not\models \mathcal{F}:$
We prove it by contradiction. Suppose that $\mathcal{F}^* \models \mathcal{F}$, then for all 
$y$ that satisfy $\mathcal{F}^*(y)$, there exists $\Phi$ such that $\Phi \models y$ and $\Phi \subseteq K_w$. Let $\Phi \subseteq K_w / K$. Since $\mathcal{F}^* \models \mathcal{F}$, then $\Phi \subseteq K$. However, $\Phi \subseteq K_w / K$, it implies that $\Phi = \emptyset$, but $\Phi \neq \emptyset$, leading to a contradiction, thus the conclusion is not valid.

\end{proof}

\section{Appendix B Other candidate functions}
Arithmetic mean: 
\begin{equation}
\mathcal F_{t}= \frac{1}{2}* ( p_{max} + \frac{1}{\eta\cdot\mathcal{H}_t + 1} ) - \gamma.
\label{eq: m1}
\end{equation}
Harmonic mean:
\begin{equation}
\mathcal F_{t}= \frac{2 \cdot p_{max}}{1+p^{max}({\eta\cdot\mathcal{H}_t + 1})} - \gamma.
\label{eq: m2}
\end{equation}

\begin{table}[]
\centering
\scalebox{0.88}{
\begin{tabular}{l|cccccc}
\toprule
\textbf{}      & \multicolumn{3}{c}{\textbf{WoW(unseen)}}                 & \multicolumn{3}{c}{\textsc{FaithDial}} \\ \cmidrule(r){2-4} \cmidrule(l){5-7}
\multicolumn{1}{l|}{} & \textbf{Nat.} & \textbf{Inf.} & \textbf{Coh.} & \textbf{Nat.} & \textbf{Inf.} & \textbf{Coh.} \\ \hline
Greedy        & 2.582 & 2.261	& 2.706 & 2.268     & 2.070     & 2.397    \\
Nucleus &  2.453  &  1.913    & 2.505   & 2.092 & 1.936  & 2.138    \\
F-Nucleus      & 2.512 & 2.159 & 2.635 &    2.236 & 1.818 & 2.414  \\
FECS     & 1.991 & 2.378 & 2.340 &    1.820 & 2.147 & 2.165  \\
DoGe  & \textbf{2.629} & \textbf{2.427} & \textbf{2.764} & \textbf{2.348} & \textbf{2.265} & \textbf{2.537}     \\ \bottomrule
\end{tabular}
}
\caption{LLMs-based Evaluation results on Wizard of Wikipedia and \textsc{FaithDial} dataset.}
\label{tab: G_eval}
\end{table}

\begin{table}[]
\centering
\renewcommand\arraystretch{1.1}
\scalebox{1.1}{
\begin{tabular}{c|c|c|c|c}
\toprule 
\textbf{Method} & Greedy & Beam & Nucleus & CD \\ \hline
\textbf{Speed} & 1.02      & 1.58    & 1.04       & 1.57         \\ \hline
\textbf{Method} & DoLa   & CS   & FECS    & DoGe      \\ \hline
\textbf{Speed} & 1.36      & 2.01    & 2.15       & 1.61        
  \\ 
\bottomrule
\end{tabular}
}
\caption{The averaged decoding speed (sec) per instance using different decoding methods on the \textsc{FaithDial} dataset. }
\label{tabspeed}
\end{table}

\begin{algorithm}[tb]
\caption{DoGe algorithm}
\label{alg:algorithm}
\textbf{Input}:   dialogue context $x$, related knowledge $k$, large language model $\mathcal{M}$, max new tokens  $N$.\\
\textbf{Parameter}: $\theta$\\
\textbf{Output}: response $y$.
\begin{algorithmic}[1] 
\STATE Let $t=0$.
\WHILE{$y_{t-1}\neq$ $<$eos$>$ and $|y|<N$}
\STATE Get $p_c(y_t) $  from $\mathcal{M}(x,y_{<t})$
\STATE Get $p_k(y_t) $  from $\mathcal{M}(x,k,y_{<t})$
\STATE Calculate $\mathcal F^c_{t} $  through Equation \ref{eq: m0}
\STATE Calculate $\mathcal F^k_{t}$  through Equation \ref{eq: m0}
\IF {$F^c_{t} > 0 \, or \, F^c_{t} > F^k_{t}$}
\STATE Calculate $\hat{p}_c(y_t) $through Equation \ref{eq: 24}
\STATE $y_t^c=$ sample$(\hat{p}_c(y_t)) $
\ELSE
\STATE Calculate $\hat{p}_k(y_t) $  through Equation \ref{eq: 255}
\STATE $y_t^k=$ argmax$(\hat{p}_k(y_t)) $
\ENDIF
\STATE $t=t+1$
\ENDWHILE
\STATE \textbf{return} $y$
\end{algorithmic}
\end{algorithm}

\section{Appendix C Baselines and hyper-parameters}
For Beam Search, we set the beam size to 3. \\
\textbf{Nucleus:} \citet{hesampling} proposed Nucleus Sampling by truncating the
unreliable tail of the probability distribution, sampling from the dynamic nucleus
of tokens containing the vast majority of the probability mass. We set the $p = 0.9$. \\
\textbf{F-Nucleus:} \citet{lee2023factuality} modified Nucleus Sampling by adapting the randomness dynamically to improve the factuality of generation. We set $p|\lambda|\omega=0.9|0.9|0.7$. \\
 \textbf{DoLa:} \citet{chuang2023dola} amplify the factual knowledge in LLM by contrasting the logits from different layers. \\
 \textbf{CS:} \citet{su2022contrastive} penalized previously generated tokens to overcome degeneration and enhance content diversity. We set $k|\alpha=3|0.6$. \\
\textbf{FECS:} \citet{chen-etal-2023-fidelity} extended Contrastive Search by integrating a faithfulness term that encourages factuality. We set $k|\alpha|\beta=3|0.3|0.3$. \\
\textbf{CD:} \citet{li2023contrastivedecodingopenendedtext} maximizes the difference
between expert log-probabilities and amateur log-probabilities, subject to plausibility constraints which restrict the search space to tokens with sufficiently high probability under the expert LM. We take the Sheared-LLaMA-1.3B \cite{xia2024shearedllamaacceleratinglanguage} as the amateur model.

\section{Appendix D Knowledge Usage Visualization}
\begin{figure}[htbp]
  \centerline{\includegraphics[scale=0.24]{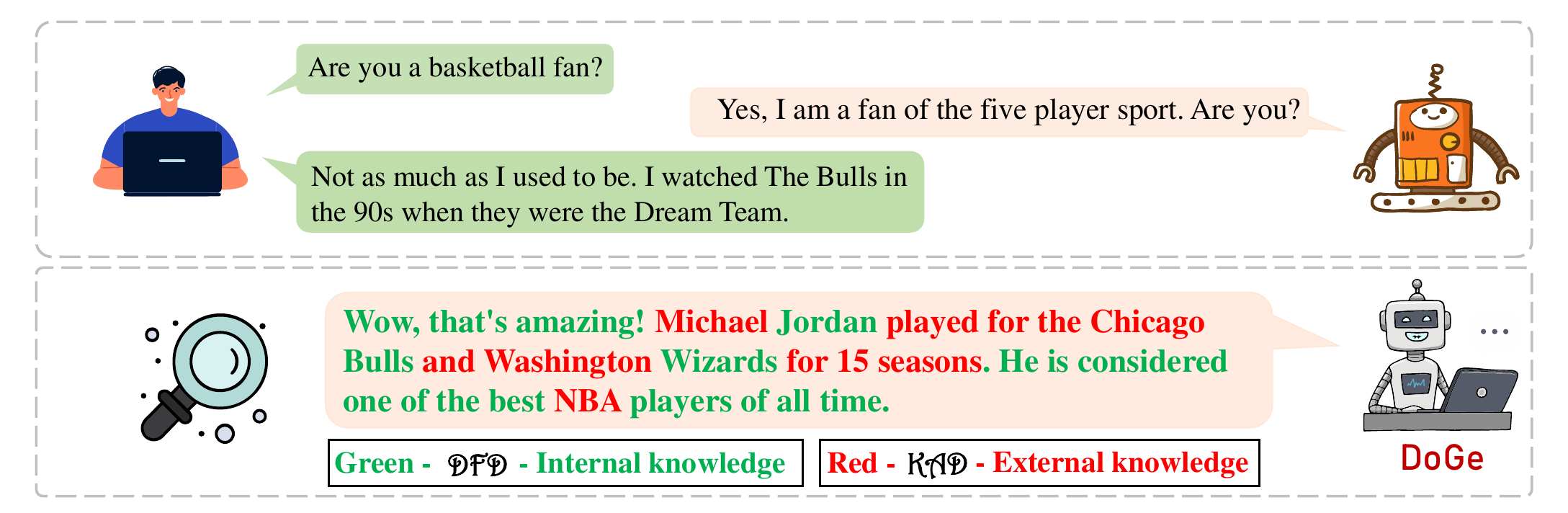}}
  \caption{A visualization of the internal and external knowledge usage for the case.}
  \label{fig-case}
\end{figure}

\section{Appendix E Details of the Prompts}
\label{prompt}
Our instruction template and demonstrations for prompting Large Language Models to generate response in knowledge-grounded dialogue generation are as in Figure \ref{Geval_template}.

\begin{table*}[]
\centering
\renewcommand\arraystretch{1.2}
\scalebox{0.67}{
\begin{tabular}{l|cccccccc|cccccccc}
\hline
\multirow{2}{*}{\textbf{Method}} & \multicolumn{7}{c}{\textbf{Wizard of Wikipedia (seen)}}                                                                              &      & \multicolumn{7}{c}{\textbf{Wizard of Wikipedia (unseen)}}                                                                           &               \\ \cline{2-17} 
                                 & \textbf{BLEU-1/2}  & \textbf{MET} & \textbf{Dist-1/2}   & \textbf{Ent-1/2}            & \textbf{BS} & \textbf{Critic}         & \textbf{Fact.}         & \textbf{CFD}  & \textbf{BLEU-1/2}  & \textbf{MET} & \textbf{Dist-1/2}   &\textbf{Ent-1/2}           & \textbf{BS} &\textbf{Critic}         & \textbf{Fact.}         &  \textbf{CFD}           \\ \hline
Greedy                          
& 31.1/20.5          & 18.5            & 8.22/43.43          & 6.59/9.76          & 65.0               & 36.33          & 88.0          & 39.72 
& 31.3/20.9          & 19.0              & 5.83/29.33          & 6.34/9.21         & 65.9             & 38.17          & 87.0          & 33.46          \\
Beam                            
& 32.2/\textbf{22.4}          & 19.8            & 8.77/45.45          & 6.78/9.95          & 74.1             & \textbf{55.02} & {90.5} & 50.01 
& {32.3/21.8} & 20.6            & 5.96/28.58          & 6.48/9.28         & 74.1             & 54.86          & \textbf{91.0} & {39.60} \\
Nucleus                          
& 26.4/15.9          & 16.2            & 7.96/47.24          & 6.64/9.99          & 57.7             & 26.85          & 84.0          & 35.61 
& 26.8/16.5          & 16.8            & 6.07/\textbf{36.25}          & 6.43/\textbf{9.62}         & 58.7             & 27.41          & 83.5          & 31.78          \\
F-Nucleus                        
& 28.0/17.5            & 16.9            & 7.98/45.64          & 6.60/9.89           & 59.9             & 29.24          & 87.0          & 36.53 
& 28.0/17.8            & 17.4            & 6.02/34.21          & 6.40/9.49          & 60.3             & 30.13          & 87.5          & 32.11          \\
DoLa                             
& 31.3/20.6          & 18.4            & 8.25/43.44          & 6.59/9.75          & 64.8             & 36.38          & 88.0          & 39.75 
& 31.6/21.1          & 18.8            & 5.89/29.45          & 6.33/9.19         & 65.5             & 38.12          & 87.5          & 33.51          \\
CD                               
& 25.8/14.4          & 17.3            & 9.33/\textbf{54.76}          & 6.77/10.10          & 63.1             & 25.78          & 82.0          & 37.57 
& 27.9/17.5          & 17.8            & 6.57/{34.81}          & 6.50/9.39          & 54.2             & 29.73          & 81.5          & 32.16          \\
CS                               
& 30.6/20.1          & 17.8            & 8.74/47.29          & 6.78/9.97          & 67.3             & 34.92          & 87.5          & 40.64 
& 29.6/19.4          & 18.0              & 6.44/32.19          & 6.43/9.37         & 63.8             & 41.59          & 88.0          & 36.59          \\
FECS                             
& 31.9/21.5          & 19.5            & 9.08/46.37          & 6.74/9.91          & 71.6             & 50.39          & 88.5          & 48.34 
& 32.5/22.0          & 19.8            & 6.35/29.84          & 6.44/9.23         & 71.5             & 51.15          & 88.5          & 39.07          \\
\textbf{DoGe}               
& \textbf{32.7*/22.4} & \textbf{20.1*}   & \textbf{9.68*}/50.07 & \textbf{6.90*/10.13} & \textbf{74.7*}    & 54.17          & \textbf{91.0} & \textbf{52.34*}
& \textbf{32.5/22.2} & \textbf{20.1}   & \textbf{6.85*}/32.95 & \textbf{6.60*}/9.46 & \textbf{75.0*}      & \textbf{55.68*} & \textbf{91.0} & \textbf{42.83*}          \\ \hline
\end{tabular}
}
\caption{Automatic Evaluation results on the WoW dataset (LLaMA2-chat-13B). 
The best results are highlighted with \textbf{bold}. "*" denotes that the improvement to the best baseline is statistically significant (t-test with $p$-value $<$  0.01). }
\label{tab: auto3}
\end{table*}

\begin{table*}[]
\centering
\renewcommand\arraystretch{1.2}
\scalebox{0.67}{
\begin{tabular}{l|cccccccc|cccccccc}
\toprule
\multirow{2}{*}{\textbf{Method}} & \multicolumn{7}{c}{\textsc{FaithDial}}                                   &      & \multicolumn{7}{c}{\textbf{Holl-E}}                                                                 &               \\ \cline{2-17} 
                                & \textbf{BLEU-1/2}  & \textbf{MET} & \textbf{Dist-1/2}   & \textbf{Ent-1/2}            & \textbf{BS} & \textbf{Critic}         & \textbf{Fact.}         & \textbf{CFD}  & \textbf{BLEU-1/2}  & \textbf{MET} & \textbf{Dist-1/2} & \textbf{Ent-1/2}            & \textbf{BS} & \textbf{Critic}         &  \textbf{Fact.}         & \textbf{CFD}         \\ \hline
Greedy                           
& 33.3/21.5                             & 21.2                                & 8.46/41.05                            & 6.54/9.54                   & 67.5                                 & 43.43     & 90.5          & 42.22 
& 47.8/40.7                             & 26.5                                & 6.20/32.09                             & 6.12/9.10                    & 62.7                                 & 34.33       & 95.5          & 33.19          \\
Beam                             
& 34.5/\textbf{23.0}                               & 22.1                                & 8.77/41.77                            & 6.66/9.67                   & 75.5                                 & 56.78   & {91.0} & 48.70 
& \textbf{57.3/52.8}                    & \textbf{34.3}                                & 6.60/34.09                             & 6.36/9.40                    & 66.7                                 & 40.32      & \textbf{96.0} & {37.07} \\
Nucleus                          
& 28.6/17.2                             & 18.6                                & 8.35/45.56                            & 6.57/9.78                   & 60.4                                 & 30.77     & 83.5         & 37.44 
& 34.3/25.9                             & 20.3                                & 6.03/36.79                            & 6.26/9.52                   & 54.8                                 & 19.77    & 91.5          & 26.97          \\
F-Nucleus                        
& 29.9/18.3                             & 19.4                                & 8.30/43.88                             & 6.54/9.70                    & 62.3                                 & 34.39    & 86.0          & 38.85 
& 38.4/30.4                             & 22.4                                & 6.17/35.65                            & 6.23/9.41                   & 57.0                                   & 23.08    & 93.5          & 28.68          \\
DoLa                             
& 33.6/21.6                             & 21.0                                  & 8.54/41.15                            & 6.53/9.52                   & 67.1                                 & 43.55        & 90.5          & 42.33 
& 47.9/40.8                             & 26.2                                & 6.23/32.02                            & 6.11/9.08                   & 62.6                                 & 34.33    & 95.0          & 33.15          \\
CD                               
& 29.3/17.6                             & 18.8                                & 9.41/\textbf{47.03}                            & 6.63/9.81                   & 58.9                                 & 30.12     & 80.5          & 37.64 
& 32.9/26.1                             & 20.4                                & 6.63/\textbf{38.08}                            & 6.39/9.53                   & 53.1                                 & 20.38     & 91.0          & 27.86          \\
CS                               
& 31.5/19.8                             & 20.1                                & 8.98/43.29                            & 6.60/9.65                    & 64.8                                 & 42.68     & 87.0          & 42.98 
& 42.6/35.6                             & 25.0                                  & 6.50/34.20                              & 6.33/9.37                   & 60.3                                 & 30.52   & 92.5          & 32.31          \\
FECS                             
& 34.9/22.8                             & 22.1                                & 9.29/43.39                            & 6.66/9.64                   & 73.4                                 & 53.16    & 88.5          & 48.03
& 50.2/44.3                               & 29.7                                & 6.59/33.65                            & 6.27/9.27                   & 65.8                                 & \textbf{40.50}   & 94.5          & 36.92          \\
\textbf{DoGe}              
& \textbf{35.0*}/22.7                      & \textbf{22.8}                       & \textbf{9.86*}/46.80                    & \textbf{6.80*/9.84}           & \textbf{76.9*}                        & \textbf{57.11}                      & 
    \textbf{92.5} &  \textbf{51.70} 
& {50.7/44.6}                    & {31.0}                         & \textbf{6.95*}/37.17                   & \textbf{6.50*/9.57*}           & \textbf{67.0}                          & {37.62} 
& \textbf{96.0} & \textbf{37.39*}          \\ \hline

\end{tabular}
}
\caption{Automatic Evaluation results on the \textsc{FaithDial} and {Holl-E} dataset (LLaMA2-chat-13B). 
The best results are highlighted with \textbf{bold}. "*" denotes that the improvement to the best baseline is statistically significant (t-test with $p$-value $<$ 0.01). }
\label{tab: auto4}
\end{table*}

\begin{figure*}[htbp]
  \centerline{\includegraphics[scale=0.23]{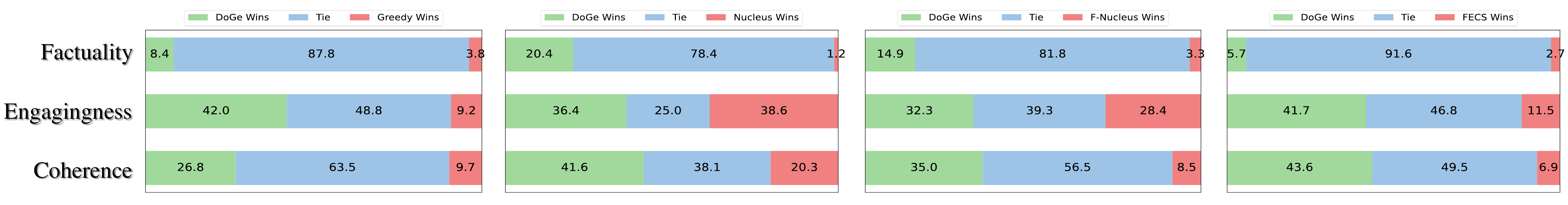}}
  \caption{Human evaluation results on the Wizard of Wikipedia test seen set. The results are statistically significant with p-value $<$ 0.05, and \textbf{Kappa} ($\kappa$) falls between 0.4 and 0.6, suggesting moderate agreement among annotators.}
  \label{fighuman}
\end{figure*}

\definecolor{awesome}{rgb}{1.0, 0.13, 0.32}
\definecolor{azure(colorwheel)}{rgb}{0.0, 0.5, 1.0}
\definecolor{aureolin}{rgb}{0.99, 0.93, 0.0}
\definecolor{amber}{rgb}{0.99, 0.93, 0.0}
\definecolor{frenchrose}{rgb}{0.96, 0.29, 0.54}
\definecolor{coquelicot}{rgb}{1.0, 0.22, 0.0}
\definecolor{aliceblue}{rgb}{0.9, 0.9, 0.9}

\lstdefinelanguage{prompt}{
    numbers=none,
    frame=shadowbox,
    framerule=0.5pt,
    framesep=2pt,
    breaklines=true,
    backgroundcolor=\color{aliceblue},
    basicstyle=\fontsize{9pt}{9pt}\selectfont\ttfamily,
    commentstyle=\color{cyan},
    morecomment=[l]{//},
    moredelim=[is][\color{frenchrose}\bfseries]{<<<}{>>>},
    moredelim=[is][\color{awesome}\bfseries]{***}{***},
    moredelim=[is][\color{azure(colorwheel)}\bfseries]{///}{///},
    moredelim=[is][\color{coquelicot}\bfseries]{|||}{|||},
}\begin{figure*}[!htb]\begin{lstlisting}[language=prompt]
***[Instruction]***
You are a chit-chat robot chatting with a user and you will be provided with your 
dialogue history and a piece of knowledge related to the user's last utterance. 
Understand this  knowledge and use it to generate a concise (no longer than 30 words) 
but informative (containing some attractive knowledge) reply.
The followings are some demonstrations you can use as reference.
***[Demonstration 1]***
The following is a multi-round dialogue between the user and you. 
User's utterance: Red is my favorite color.  My house, car, and clothes are all red.  
Your response: I also like red buddy, I have more red colour dresses.  User's last 
utterance: What do you think it is about red that makes it so appealing?  Related 
knowledge: Red is the colour at the end of the visible spectrum of light, next to orange 
and opposite violet. Your knowledge-grounded response to the User's last utterance: Red 
is visible of light, it is next to orange and opposite to violet.
***[Demonstration 2]***
The following is a multi-round dialogue between the user and you. 
User's utterance: Is rock and roll still popular today?  Your response: It's hard to say. 
However, radio stations have much success playing classic rock and roll, which is a sub 
genre that usually has one or two electric guitars, a double bass or string bass or 
electric bass guitar, and a drum kit.  User's utterance: I used to listen to the rock 
band Rolling Stones.  Are they still around today?  Your response: They are! Even though 
they were formed in 1962 and have had a long list of line-up changes, they're still 
around today, with Mick Jagger still leading the band.  User's last utterance: Wow, that 
is a long time to be playing music.  I wonder if any other bands have been around that 
long.  Related knowledge: Red Hot Chili Peppers are an American funk rock band formed in 
Los Angeles in 1983. Your knowledge-grounded response to the User's last utterance: It 
all depends! You have bands like the Red Hot Chili Peppers who, although have not reached 
the popularity of the Rolling Stones, have been around since 1983 themselves.
***[Demonstration 3]***
The following is a multi-round dialogue between the user and you. User's utterance: Gouda 
cheese  Your response: Do you know that Gouda cheese is made from cow milk? User's last 
utterance: Most all cheese if made from Cow milk I think. I'm a fan of Gouda, it is 
pretty good. Related knowledge: It is one of the most popular cheeses worldwide. Your 
knowledge-grounded response to the User's last utterance: I bet because it is one of the 
most popular cheeses worldwide.
***[Target conversation]***
Now complete the following dialogue:
User's utterance: |||{Dialogue History}|||
User's last utterance: |||{User's Query}|||
Related knowledge: |||{External Knowledge}|||
Your knowledge-grounded response to the User's last utterance: 

\end{lstlisting}
\caption{Instruction template and demonstrations for prompting the LLMs.}
\label{Geval_template}
\end{figure*}

\end{document}